\documentclass[10pt,twocolumn,letterpaper]{article}

\usepackage[pagenumbers]{cvpr} 

\usepackage{graphicx}
\usepackage{amsmath}
\usepackage{amssymb}
\usepackage{booktabs}
\usepackage{adjustbox}
\usepackage{floatrow}
\usepackage{multirow}
\usepackage{subcaption}
\usepackage{pifont}
\usepackage{floatrow}
\usepackage{lipsum}

\newcommand{\cmark}{\ding{51}}%
\newcommand{\xmark}{\ding{55}}%

\usepackage[pagebackref,breaklinks,colorlinks]{hyperref}

\usepackage[capitalize]{cleveref}
\crefname{section}{Sec.}{Secs.}
\Crefname{section}{Section}{Sections}
\Crefname{table}{Table}{Tables}
\crefname{table}{Tab.}{Tabs.}

\def\cvprPaperID{730} 
\def\confName{CVPR}
\def\confYear{2022}

\begin{document}

\title{MS-TCT: Multi-Scale Temporal ConvTransformer for Action Detection}
\author{\large
Rui Dai\textsuperscript{1,2},\hskip 1em
Srijan Das\textsuperscript{3},\hskip 1em
Kumara Kahatapitiya\textsuperscript{3}, \hskip 1em 
Michael S. Ryoo\textsuperscript{3},\hskip 1em
Fran\c cois Brémond\textsuperscript{1,2}
\\
\textsuperscript{1}Inria \hskip 1em
\textsuperscript{2}Université Côte d'Azur \hskip 1em
\textsuperscript{3}Stony Brook University \hskip 1em
\\
{\tt\small \textsuperscript{1,2}\{name.surname\}@inria.fr} 
}
\maketitle

\begin{abstract}
   Action detection is a significant and challenging task, especially in densely-labelled datasets of untrimmed videos. Such data consist of complex temporal relations including composite or co-occurring actions. To detect actions in these complex settings, it is critical to capture both short-term and long-term temporal information efficiently. To this end, we propose a novel `ConvTransformer' network for action detection: MS-TCT\footnote{Code/ Models: \url{https://github.com/dairui01/MS-TCT}}. This network comprises of three main components: (1) a Temporal Encoder module which explores global and local temporal relations at multiple temporal resolutions, (2) a Temporal Scale Mixer module which effectively fuses multi-scale features, creating a unified feature representation, and (3) a Classification module which learns a center-relative position of each action instance in time, and predicts frame-level classification scores. Our experimental results on multiple challenging datasets such as Charades, TSU and MultiTHUMOS, validate the effectiveness of the proposed method, which outperforms the state-of-the-art methods on all three datasets. 
\end{abstract}

\vspace{-5mm}
\section{Introduction}
\label{sec:intro}
\begin{figure}[t]
\centering
\includegraphics[width=8.2cm]{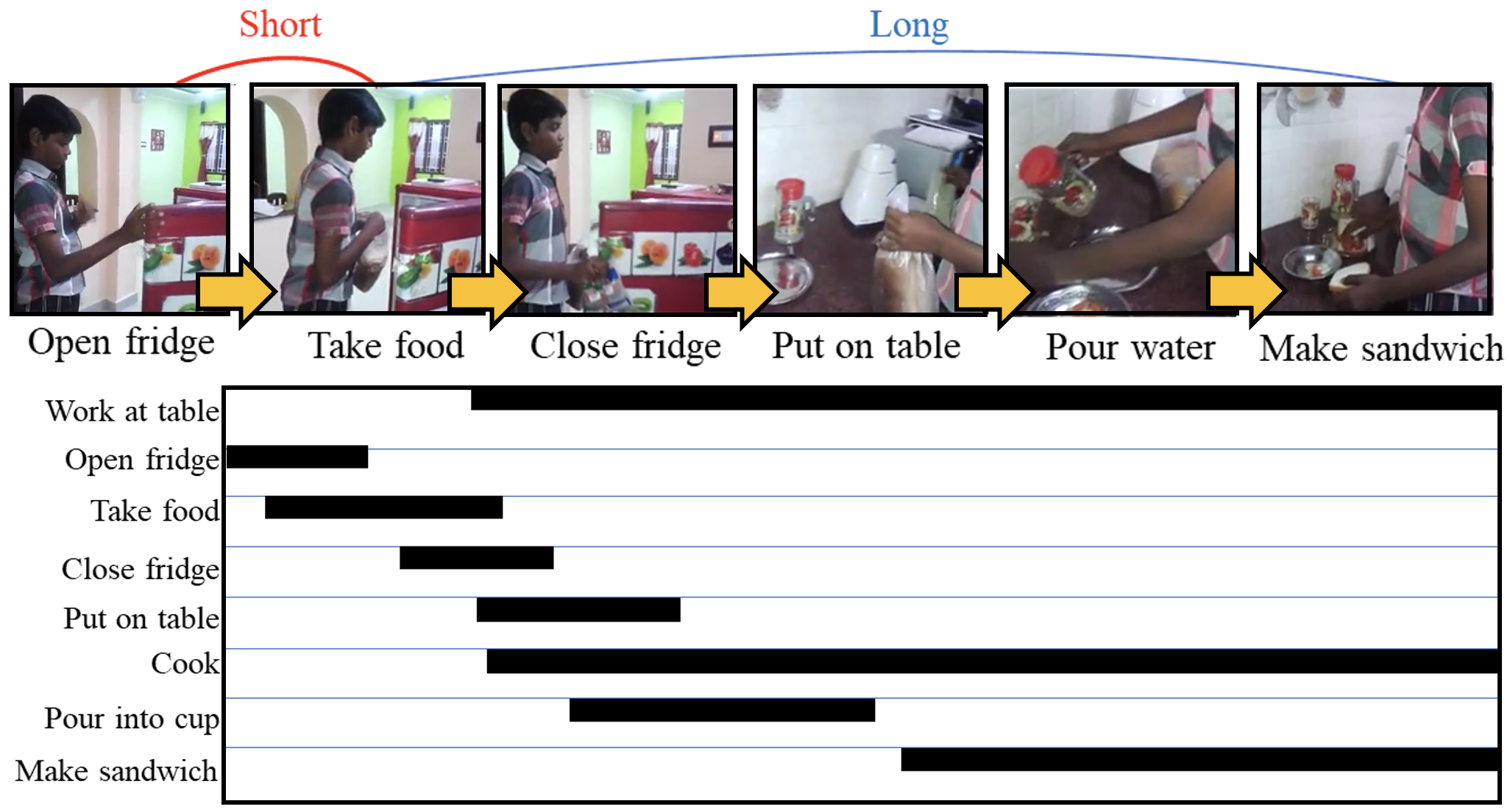}
\caption{\textbf{Complex temporal relations in untrimmed videos:} Here, we show a typical distribution of actions in a densely-labelled video, which consists of both long-term and short-term dependencies among actions. 
}
\label{fig:front}
\end{figure}

Action detection is a well-known problem in computer vision, which is aimed towards finding precise temporal boundaries among actions occurring in untrimmed videos. 
It aligns well with real-world settings, because every minute of a video is potentially filled with multiple actions to be detected and labelled. 
There are public datasets~\cite{charades,dai2020toyota,multi-thumos} which provide dense annotations to tackle this problem, having an action distribution similar to the real-world. 
However, such data can be challenging, with multiple actions occurring concurrently over different time spans, and having limited background information. Therefore, understanding both short-term and long-term temporal dependencies among actions is critical for making good predictions.
For instance, the action of \textit{`taking food'} (see Fig.~\ref{fig:front}) can get context information from \textit{`opening fridge'} and \textit{`making sandwich'}, which correspond to the short-term and long-term action dependencies, respectively. Also, the occurrence of \textit{`putting something on the table'} and \textit{`making sandwich'} provide contextual information to detect the composite action \textit{`cooking'}. 
This example shows the need for an effective temporal modeling technique for detecting actions in a densely-labelled videos.

Towards modeling temporal relations in untrimmed videos, multiple previous methods~\cite{Dai_2021_WACV,dai2021ctrn,lea2017temporal,dai2019self,dai2019tan,TGM1} use 1D temporal convolutions~\cite{lea2017temporal}. However, limited by 
their kernel size, convolution-based methods can directly access local information only, not learning direct relations between temporally-distant segments in a video (here, we consider a set of consecutive frames as a segment). Thus, such methods fail to model long-range interactions between segments which may be important for action detection.
With the success of Transformers~\cite{transformer,zhu2020deformable,dosovitskiy2020image,liu2021swin} in natural language processing and more recently in computer vision, recent methods~\cite{MLAD,tan2021relaxed} have leveraged multi-head self-attention (MHSA) to model long-term relations in videos for action detection.
Such attention mechanisms can build direct one-to-one global relationships between each temporal segment (i.e., temporal token) of a video to detect highly-correlated and composite actions. However, existing methods rely on modeling such long-term relationships on input frames themselves. Here, a temporal token covers only a few frames, which is often too short w.r.t.~to the duration of action instances. Also, in this setting, transformers need to explicitly learn strong relationships between adjacent tokens which arise due to temporal consistency, whereas it comes naturally for temporal convolutions (i.e., local inductive bias).
Therefore, a pure transformer architecture
may not be sufficient to model complex temporal dependencies for action detection.

\begin{figure}[t]
  \centering
  \includegraphics[width=\linewidth]{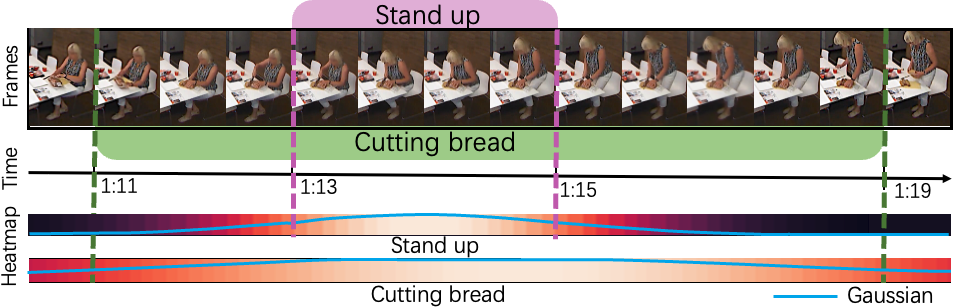}
  \caption{\textbf{Relative temporal position heat-map ($\mathbf{G^*}$):} We present a video clip which contains two overlapping action instances. The \textit{Gaussians} indicate the intensities of temporal heat-maps, which are centered at the mid point of each action, in time.}
  \label{fig:relative}
\end{figure}

\noindent To this end, we propose \textit{\textbf{Multi-Scale Temporal ConvTransformer (MS-TCT)}}, a model which benefits from both convolutions and self-attention. We use convolutions in a token-based architecture to promote multiple temporal scales of tokens, and to blend neighboring tokens imposing a temporal consistency with ease.
In fact, MS-TCT is built on top of temporal segments encoded using a 3D convolutional backbone~\cite{i3d}. Each temporal segment is considered as a single input token to MS-TCT, to be processed in multiple stages with different temporal scales. These scales are determined by the size of the temporal segment, which is considered as a single token
at the input of each stage. Having different scales allows MS-TCT to learn both fine-grained relations between atomic actions (e.g. \textit{`open fridge'}) in the early stages, and coarse relations between composite actions (e.g. \textit{`cooking'}) in the latter stages. 
To be more specific, each stage consists of a temporal convolution layer for merging tokens, followed by a set of multi-head self-attention layers and temporal convolution layers, which model global temporal relations and infuse local information among tokens, respectively. 
%
%
As convolution introduces an inductive bias~\cite{d2021convit}, the use of temporal convolution layers in MS-TCT can infuse positional information related to tokens~\cite{guo2021cmt,islam2020much}, even without having any positional embeddings, unlike pure transformers~\cite{dosovitskiy2020image}. 
Followed by the modeling of temporal relations at different scales, a mixer module is used to fuse the features from each stages to get a unified feature representation. 
Finally, to predict densely-distributed actions, we introduce a heat-map branch in MS-TCT in addition to the usual multi-label classification branch. This heat-map encourages the network to predict the relative temporal position of instances of each action class. 
Fig.~\ref{fig:relative} shows the relative temporal position, which is computed based on a Gaussian filter parameterized by the instance center and its duration. 
It represents the relative temporal position w.r.t.~to the action instance center at any given time. 
With this new branch, MS-TCT can embed a class-wise relative temporal position in token representations, encouraging discriminative token classification in complex videos. 

%

To summarize, the main contributions of this work are to
(1) propose an effective and efficient ConvTransformer for modeling complex temporal relations in untrimmed videos, (2) introduce a new branch to learn the position relative to instance-center, which promotes action detection in densely-labelled videos, and (3) improve the state-of-the-art on three challenging densely-labelled action datasets. 

\section{Related Work}
\label{sec:relate}
Action detection has received a lot of interest in recent years~\cite{gleason2019proposal,lin2017single,gtad,Damen2018EPICKITCHENS,dai2019tan,zhao2019hacs,Dai_2021_ICCV}. In this work, we focus on action detection in densely-labelled videos~\cite{charades,dai2020toyota,multi-thumos}. 
The early attempts on modeling complex temporal relations tend to use anchor-based methods~\cite{RC3d,AFNET}. However, dense action distributions require large amount of such anchors. 
%
Superevent~\cite{superevent} utilizes a set of Gaussian filters to learn video glimpses, which are later summed up with a soft attention mechanism to form a global representation. 
However, as these Gaussians are independent of the input videos, it can not handle videos with minor frequencies of composite actions effectively.
Similarly, TGM~\cite{TGM1} is also a temporal filter based on Gaussian distributions, which enables the learning of longer temporal structures with a limited number of parameters. 
PDAN~\cite{Dai_2021_WACV} is a temporal convolutional network, with temporal kernels which are adaptive to the input data. Although TGM and PDAN achieve state-of-the-art performance in modeling complex temporal relations, these relations are constrained to local regions, thus preventing them to learn long-range relationships. 
Coarse-Fine Networks~\cite{kahatapitiya2021coarse} leverage two X3D~\cite{x3d} networks in a Slow-Fast~\cite{slow_fast} fashion. This network can jointly model spatio-temporal relations. However, it is limited by the number of input frames in X3D backbone, and a large stride is required to process long videos efficiently. This prevents Coarse-Fine Networks from considering the fine-grained details in long videos for detecting action boundaries. A concurrent work \cite{kahatapitiya2021self} looks into detection pretraining with only classification labels, to improve downstream action detection.
Recently, some attempts have been proposed to model long-term relationships explicitly: MTCN~\cite{kazakos2021MTCN} benefits from the temporal context of action and labels, whereas TQN~\cite{zhangtqn} factorizes categories into pre-defined attribute queries to predict fine-grained actions. However, it is not trivial to extend both approaches to action detection in untrimmed videos. 

\begin{figure*}[t!]
\centering
\includegraphics[width=15cm]{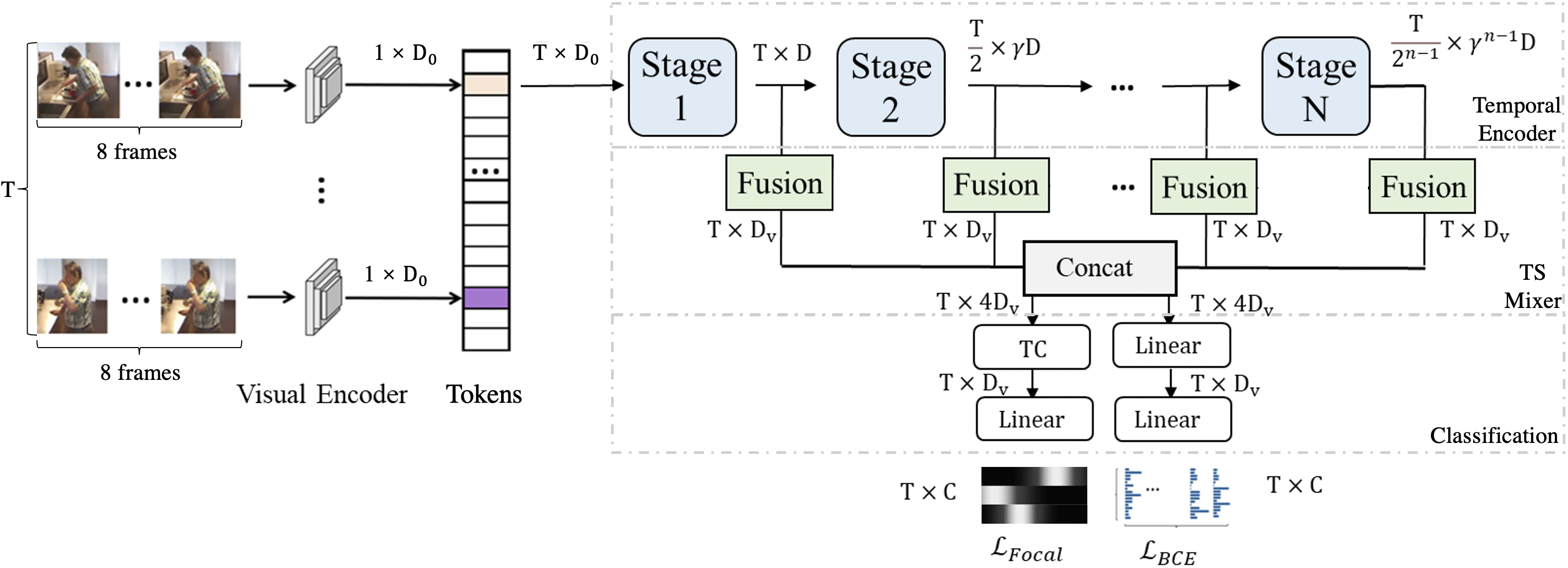}\vspace{-0.1in}
\caption{\textbf{Multi-Scale Temporal ConvTransformer (MS-TCT)} for action detection consists of four main components: (1) a Visual Encoder, (2) a Temporal Encoder, 
(3) a Temporal Scale Mixer (TS Mixer) and 
(4) a Classification Module. Here, $TC$ indicates the 1D convolutional layer with kernel size $k$. 
}
\label{fig:main}
\end{figure*}


Recent Transformer models have been successful in both image and video domain~\cite{zhu2020deformable,dosovitskiy2020image,liu2021swin,zhou2021deepvit,wang2021pyramid,wang2021pvtv2,xie2021segformer,cheng2021per,arnab2021vivit,liu2021video,timesformer, ryoo2021tokenlearner}.
Although Vision Transformers such as TimeSformer~\cite{transformer} can consider frame-level input tokens to model temporal relations, it is limited to short video clips which is insufficient to model fine-grained details in longer real-world videos. 
As a compromise, recent action detection methods use multi-head self-attention layers on top of the visual segments encoded by 3D convolutional backbones~\cite{i3d}. 
RTD-Net~\cite{tan2021relaxed}, an extension of DETR~\cite{zhu2020deformable}, uses a transformer decoder to model the relations between the proposal and the tokens. However, this network is designed only for sparsely-annotated videos~\cite{THUMOS14,caba2015activitynet}, where only a single action exists per video. In dense action distributions, the module that detects the boundaries in RTD-Net fails to separate foreground and background regions. 
MLAD~\cite{MLAD} learns class-specific features and uses a transformer encoder to model class relations at each time-step and temporal relations for each class. 
However, MLAD struggles with datasets that has complex labels~\cite{charades}, since it is hard to extract class-specific features in such videos.
In contrast to these transformers introduced for action detection, we propose a ConvTransformer: \textit{MS-TCT}, which inherits a transformer encoder architecture, while also gaining benefits from temporal convolution. Our method can model temporal tokens both globally and locally at different temporal scales. 
Although other ConvTransformers~\cite{wu2021cvt,d2021convit,guo2021cmt, kahatapitiya2021swat} exist for image classification,
our network is designed and rooted for densely-labelled action detection.


\section{Multi-Scale Temporal ConvTransformer}
\label{sec:method}

First, we define the problem statement of action detection in densely-labelled settings. Formally, for a video sequence of length $T$, each time-step $t$ contains a ground-truth action label $y_{t,c} \in \{0, 1\}$, where $c \in \{1,...,C\}$ indicates an action class. For each time-step, an action detection model needs to predict class probabilities $\tilde{y}_{t,c} \in [0, 1]$.
Here, we describe our proposed action detection network: \textit{MS-TCT}. 
As depicted in Fig.~\ref{fig:main}, it consists of four main components:
(1) a \textit{\textbf{Visual Encoder}} which encodes a preliminary video representation, (2) a \textit{\textbf{Temporal Encoder}} which structurally models the temporal relations at different temporal scales (i.e., resolution), (3) a \textit{\textbf{ Temporal Scale Mixer}}, dubbed as~\textit{TS Mixer}, which combines multi-scale temporal representations, and (4) a \textit{\textbf{Classification Module}} which predicts class probabilities. 
In the following sections, we present the details of each these components of MS-TCT. 

\subsection{Visual Encoder} 
%
The input to our action detection network: MS-TCT, is an untrimmed video which may span for a long duration~\cite{dai2020toyota} (e.g. multiple minutes). 
However, processing long videos in both spatial and temporal dimensions can be challenging, mainly due to computational burden. As a compromise, similar to previous action detection models~\cite{MLAD,Dai_2021_WACV,TGM1}, we consider features of video segments extracted by a 3D CNN as inputs to MS-TCT, which embed spatial information latently as channels. 
Specifically, we use an I3D backbone~\cite{i3d} to encode videos. Each video is divided into $T$ non-overlapping segments (during training), each of which consists of 8 frames. Such RGB frames are fed as an input segment to the I3D network.
Each segment-level feature (output of I3D) can be seen as a transformer token of a time-step (i.e., temporal token).
We stack the tokens along the temporal axis to form a $T\times D_0$ video token representation, to be fed in to the Temporal Encoder.

\subsection{Temporal Encoder}
\begin{figure}[ht!]
\centering
\includegraphics[width=7cm]{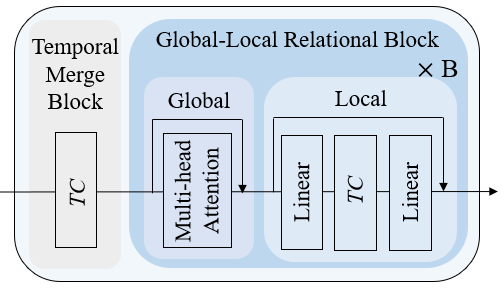}
\caption{\textbf{A single stage of our Temporal Encoder} consists of (1) a Temporal Merging Block and (2) $\times B$ Global-Local Relational Blocks. Each Global-Local Relational Block contains a Global and a Local Relational Block. Here, $Linear$ and $TC$ indicates the 1D convolutional layer with kernel size 1 and $k$ respectively.
}
\label{fig:stage}
\end{figure}
As previously highlighted in Section~\ref{sec:intro}, efficient temporal modeling is critical for understanding long-term temporal relations in a video, especially for complex action compositions. 
Given a set of video tokens, there are two main ways to model temporal information: using (1) a 1D Temporal Convolutional layer~\cite{lea2017temporal}, which focuses on the neighboring tokens but overlooks the direct long-term temporal dependencies in a video, or
(2) a Transformer~\cite{transformer} layer that globally encodes one-to-one interactions of all tokens, while neglecting the local semantics, which has proven beneficial in modeling the highly-correlated visual signals~\cite{hubel1962receptive,fukushima1975cognitron}. 
%
Our Temporal Encoder benefits from the best of both worlds, by exploring both local and global contextual information in an alternating fashion. 

As shown in Fig.~\ref{fig:main}, Temporal Encoder follows a hierarchical structure with $N$ stages: Earlier stages learn a fine-grained action representation with more temporal tokens, whereas the latter stages learn a coarse representation with fewer tokens. 
%
%
Each stage corresponds to a semantic level (i.e., temporal resolution) and consists of one Temporal Merging block and $\times B$ Global-Local Relational Blocks (see Fig.~\ref{fig:stage}): 

\noindent\textbf{Temporal Merging Block} is the key component for introducing network hierarchy, which shrinks the number of tokens (i.e.,~temporal resolution) while increasing the feature dimension. 
This step can be seen as a weighted pooling operation among the neighboring tokens. 
In practice, we use a single temporal convolutional layer (with a kernel size of $k$, and a stride of 2, in general) to halve the number of tokens and extend the channel size by $\times\gamma$. 
In the first stage, we keep a stride of 1 to maintain the same number of tokens as the I3D output, and project the feature size from $D_{0}$ to $D$ (see Fig.~\ref{fig:main}). This is simply a design choice. 

\noindent\textbf{Global-Local Relational Block} is further decomposed in to a \textit{Global Relational Block} and a \textit{Local Relational Block} (see Fig.~\ref{fig:stage}). 
In Global Relational Block, we use the standard multi-head self-attention layer~\cite{transformer} to model long-term action dependencies, i.e., global contextual relations. 
In Local Relational Block, we use a temporal convolutional layer (with a kernel size of $k$) to enhance the token representation by infusing the contextual information from the neighboring tokens, i.e., local inductive bias. This enhances the temporal consistency of each token while modeling the short-term temporal information corresponding to an action instance.
%
%
%

In the following, we formulate the computation flow inside the Global-Local Relational Block. For brevity, here, we drop the stage index $n$. 
For a block $j \in \{1,...,B\}$, we represent the input tokens as $X_j \in \mathbb{R}^{T' \times D'}$. First, the tokens go through multi-head attention layer in Global Relational Block, which consists of $H$ attention heads. 
For each head $i \in \{1,...,H\}$, an input $X_j$ is projected in to $Q_{ij}=W^Q_{ij}X_j$, $K_{ij}=W^K_{ij}X_j$ and $V_{ij}=W^V_{ij}X_j$, where $W^Q_{ij}$, $W^K_{ij}$, $W^V_{ij} \in \mathbb{R}^{D_h \times D'}$ represent the weights of linear layers and $D_h = \frac{D'}{H}$ represents the feature dimension of each head. 
Consequently, the self-attention for head $i$ is computed as,
\begin{equation}
\small
Att_{ij} = \texttt{Softmax}(\frac{Q_{ij} K_{ij}^\top}{\sqrt{D_{h}}})V_{ij}\;.
\end{equation}
Then, the output of different attention heads are mixed with an additional linear layer as,
\begin{equation}
\small
    M_j = W_j^O \texttt{Concat}(Att_{1j}, ..., Att_{Hj}) + X_j\;,
\end{equation}
where $W^O_j \in \mathbb{R}^{D' \times D'}$ represents the weight of the linear layer. The output feature size of multi-head attention layer is the same as the input feature size.

Next, the output tokens of multi-head attention are fed in to the \textit{Local Relational Block}, which consists of two linear layers and a temporal convolutional layer. 
As shown in Fig.~\ref{fig:stage}, the tokens first go through a linear layer to increase the feature dimension from $D'$ to $\theta D'$, followed by a temporal convolutional layer with a kernel size of $k$, which blends the neighboring tokens to provide local positional information to the temporal tokens~\cite{islam2020much}. Finally, another linear layer projects the feature dimension back to $D'$.
The two linear layers in this block enable the transition between the multi-head attention layer and temporal convolutional layer. 
The output feature dimension remains the same as the input feature for the Local Relational Block. This output is fed to the next Global Relational Block if block $j < B$.

The output tokens from the last Global-Local Relational Block from each stage are combined and fed to the following Temporal Scale Mixer.

\subsection{Temporal Scale Mixer}

After obtaining the tokens at different temporal scales, the question that remains is, \textit{how to aggregate such multi-scale tokens to have a unified video representation}?
To predict the action probabilities, our classification module needs to make predictions at the original temporal length as the network input. Thus, we require to interpolate the tokens across the temporal dimension, which is achieved by performing an up-sampling and a linear projection step. 
\begin{figure}[t!]
\centering
\includegraphics[width=5cm]{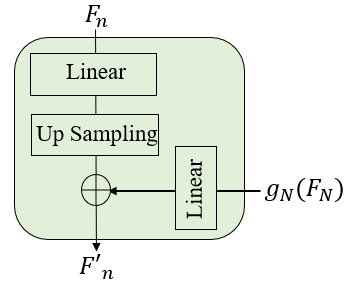}
\caption{\textbf{Temporal Scale Mixer Module:} The output tokens $F_n$ of stage $n$ is resized and up-sampled to $T\times D_v$, then summed with the tokens from the last stage $N$. 
}
\label{fig:mixer}
\end{figure}
As shown in Fig.~\ref{fig:mixer}, for the output $F_n$ from stage $n \in \{1,...,N\}$, this operation can be formulated as,
\begin{equation}
\small
g_n(F_{n}) =  \texttt{UpSampling}_n(F_{n}W^n)\;,
\end{equation}
where $W^n \in \mathbb{R}^{D_v \times \gamma^{n-1} D }$ with an upsampling rate of $n$.
In our hierarchical architecture, earlier stages (with lower semantics) have higher temporal resolution, whereas the latter stages (with high semantics) have lower temporal resolution.
To balance the resolution and semantics, upsampled tokens from the last stage $N$ is processed through a linear layer and summed with the upsampled tokens from each stage ($n < N$). 
This operation can be formulated as,
\begin{equation}
\small
F'_{n} = g_n(F_{n}) \oplus g_N(F_{N})W_n\;,
\end{equation}
where $F'_{n}$ is the refined tokens of stage $n$, $\oplus$ indicates the element-wise addition and $W_n \in \mathbb{R}^{D_v \times D_v}$.
Here, all the refined token representations have the same temporal length.   
Finally, we concatenate them to get the final multi-scale video representation $F_v \in \mathbb{R}^{T \times ND_v}$. 
\begin{equation}
\small
F_v = \texttt{Concat}(F'_{1}, ..., F'_{N-1}, F_{N})\;.
\end{equation}
Note that more complicated fusion methods~\cite{dai2021attentional,liu2018path} can be built on top of these multi-scale tokens. However, we see that the simple version described above performs the best. 

The multi-scale video representation $F_v$ is then sent to the classification module for making predictions.

\subsection{Classification Module}
Training MS-TCT is achieved by jointly learning two classification tasks. 
As mentioned in Section~\ref{sec:intro}, in this work, we introduce a new classification branch to learn a heat-map of the action instances. 
This heat-map is different from the ground truth label as it varies across time, based on the action center and duration. The objective of using such heat-map representation is to encode temporal relative positioning in the learned tokens of MS-TCT.

In order to train the heat-map branch, we first need to build the \textit{class-wise} ground-truth heat-map response $G^{*} \in [0,1]^{T\times C}$, where $C$ indicates the number of action classes. In this work, we construct $G^{*}$ by considering the maximum response of a set of one-dimensional Gaussian filters. Each Gaussian filter corresponds to an instance of action class in a video, centered at the specific action instance, in time. 
More precisely, for every temporal location $t$ the ground-truth heat-map response is formulated as,
\begin{equation}
\small
G^{*}_{c} (t)= \max_{a=1,...,A_c} \texttt{Gaussian}(t, t_{a,c}; \sigma)\;,
\end{equation}
\begin{equation}
\small 
\texttt{Gaussian}(t, t_{a,c};\sigma) = \frac{1}{\sqrt{2 \pi}\sigma} \texttt{exp}^{-\frac{(t - t_{a,c})^2}{2\sigma^2}}\;.
\end{equation}
Here, $\texttt{Gaussian}(\cdot, \cdot; \sigma)$ provides an instance-specific Gaussian activation according to the center and instance duration. Moreover, $\sigma$ is equal to $\frac{1}{2}$ of each instance duration and $t_{a,c}$ represents the center for class $c$ and instance $a$. $A_c$ is the total number of instances for class $c$ in the video.
As shown in Fig.~\ref{fig:main}, heat-map $G$ is computed using a temporal convolutional layer with a kernel size of $k$ and a non-linear activation, followed by another linear layer with a sigmoid activation. 
Given the ground-truth $G^{*}$ and the predicted heat-map $G$, we compute the \textit{action focal loss} ~\cite{zhou2019objects,lin2017focal} which is formulated as,
\begin{equation}
\small
\mathcal{L}_{\text{Focal}} = \frac{1}{A}\sum_{t,c}\begin{cases}
(1-G_{t,c})^2 \texttt{log}(G_{t,c}) \text{~~~~~~~~~~~~~~~~~~~if~~}G^{*}_{t,c}=1\;, \\
(1-G^{*}_{t,c})^{4}(G_{t,c})^2 \texttt{log}(1-G_{t,c})\text{~~Otherwise}\;,
\end{cases}
\end{equation}
where $A$ is the total number of action instances in a video. 

Similar to the previous work~\cite{Dai_2021_WACV,MLAD}, we leverage another branch to perform the usual multi-label classification. With video features $F_v$, the predictions are computed using two linear layers with a sigmoid activation, and Binary Cross Entropy (BCE) loss~\cite{nam2014large} is computed against the ground-truth labels. Only the scores predicted from this branch are used in evaluation. 
Input to both the branches are the same output tokens $F_v$. 
The heat-map branch encourages the model to embed the relative position w.r.t. the instance center in to video tokens $F_v$.  
Consequently, the classification branch can also benefit from such positional information to make better predictions.

The overall loss is formulated as a weighted sum of the two losses mentioned above, with the weight $\alpha$ is chosen according to the numerical scale of losses.
\begin{equation}
\small
\mathcal{L}_{\text{Total}} = \mathcal{L}_{\text{BCE}} + \alpha\; \mathcal{L}_{\text{Focal}}\;.
\end{equation}
\section{Experiments}
\label{sec:experiment}
\noindent\textbf{Datasets:} We evaluate our framework on three challenging multi-label action detection datasets: Charades~\cite{charades}, TSU~\cite{dai2020toyota} and MultiTHUMOS~\cite{multi-thumos}. 
Charades~\cite{charades} is a large dataset with 9848 videos of daily indoor actions. The dataset contains 66K+ temporal annotations for 157 action classes, with a high overlap among action instances of different classes.  This is in contrast to other action detection datasets such as ActivityNet~\cite{caba2015activitynet}, which only have one action per time-step. We evaluate on the localization setting of the dataset~\cite{sigurdsson2017asynchronous}. 
Similar to the Charades, TSU~\cite{dai2020toyota} is also recorded in indoor environment with dense annotations. Up to 5 actions can happen at the same time in a given frame. However, different from Charades, TSU has many long-term composite actions. 
MultiTHUMOS~\cite{multi-thumos} is an extended version of THUMOS’14~\cite{THUMOS14}, containing dense, multi-label action annotations for 65 classes across 413 sports videos. 
By default, we evaluate the per-frame mAP on these densely-labelled datasets following~\cite{sigurdsson2017asynchronous,multi-thumos}.

\noindent\textbf{Implementation Details:} In the proposed network, we use number of stage $N=4$ the number of Global-Local Relational Blocks $B=3$ for each stage. Note that for small dataset as MultiTHUMOS, $B=2$ is sufficient.
The number of attention heads for the Global Relational Block is set to 8. We use the same output feature dimension of I3D (after \texttt{Global Average Pooling} as input to MS-TCT, and thus $D_0=1024$. Input features are then projected in to $D=256$ dimensional feature using the temporal merging block in the first stage. We consider feature expansion rate $\gamma=1.5$ and $\theta=8$. Kernel size $k$ of temporal convolutional layer is set to be 3, with zero padding to maintain the resolution. The loss balance factor $\alpha=0.05$. 
The number of tokens is fixed to $T =256$ as input to MS-TCT. During training, we randomly sample consecutive $T$ tokens from a given I3D feature representation. At inference, we follow~\cite{MLAD} to use a sliding window approach to make predictions. 
Our model is trained on two GTX 1080 Ti GPUs with a batch-size of 32. We use Adam optimizer~\cite{adam_optimizer} with an initial learning rate of 0.0001, which is scaled by a factor of 0.5 with a patience of 8 epochs.

\subsection{Ablation Study}
In this section, we study the effectiveness of each component in the proposed network on Charades dataset. 

\noindent \textbf{Importance of Each Component in MS-TCT:}
As shown in Table~\ref{tab:ablation1}, I3D features with the classification branch only, is considered as the representative baseline.
This baseline consists in a classifier that discriminates the I3D features at each time-step without any further temporal modeling.  
On top of that, adding our Temporal Encoder significantly improves the performance (+ 7.0\%) \wrt I3D feature baseline. This improvement reflects the effectiveness of the Temporal Encoder in modeling the temporal relations within the videos. 
In addition, if we introduce a Temporal Scale Mixer to blend the features from different temporal scales, it gives a + 0.5\% improvement, with minimal increase in computations. 
Finally, we study the utility of our heat-map branch in the classification module. We find that the heat-map branch is effective when optimized along with the classification branch, but fails to learn discriminative representations when optimized without it (25.4\% vs 10.7\%). The heat-map branch encourages the tokens to predict the action center while down-playing the tokens towards action boundaries. In comparison, the classification branch improves the token representations equally for all tokens, despite action boundaries. Thus, when optimized together, both branches enable the model to learn a better action representation.
%
While having all the components, the proposed network achieves a significant + 9.8\% improvement \wrt I3D feature baseline validating that each component in MS-TCT is instrumental for the task of action detection.

\noindent \textbf{Design Choice for a Stage:} In Table~\ref{tab:ablation2}, we present the ablation related to the design choices of a stage in the Temporal Encoder. Each row in Table~\ref{tab:ablation2} indicates the result of removing a component in each stage. 
%
Note that, removing the Temporal Merge block indicates replacing this block with a temporal convolutional layer of stride 1, i.e., only the channel dimension is modified across stages. 
In Table~\ref{tab:ablation2}, we find that removing any component can drop the performance with a significant margin. 
This observation shows the importance of jointly modeling both global and local relations in our method, and the effectiveness of the multi-scale structure.
These properties in MS-TCT make it easier to learn complex temporal relationships which span across both (1) neighboring temporal segments, and (2) distant temporal segments.

\begin{table}[t]
\caption{\textbf{Ablation on each component in MS-TCT:} The evaluation is based on per-frame mAP on Charades dataset.}
\centering
\label{tab:ablation1}
\scalebox{0.9}{
\begin{tabular}{cccc|c}
\hline
Temporal  & TS  & Heat-Map  & Classification & mAP  \\
Encoder & Mixer & Branch & Branch & (\%) \\\hline
\xmark                   & \xmark             & \xmark   & \cmark     & 15.6 \\
\cmark                   & \xmark             & \xmark    & \cmark   & 23.6 \\
\cmark                   & \cmark             & \xmark    & \cmark   & 24.1 \\
\cmark                   & \cmark             & \cmark    & \xmark    & 10.7 \\\hline
\cmark                   & \cmark             & \cmark    & \cmark    & \textbf{25.4} \\
\hline
\end{tabular}}
\end{table}

\begin{table}[t]\tabcolsep=12pt
\caption{\textbf{Ablation on the design of a single stage in our Temporal Encoder}, evaluated using per-frame mAP on Charades dataset.}
\centering
\label{tab:ablation2}
\scalebox{0.9}{
\begin{tabular}{ccc|c}
\hline
Temporal &Global & Local & mAP \\
Merge &Layer & Layer & (\%) \\\hline
\cmark&\cmark       & \xmark  &24.0      \\
\cmark& \xmark       & \cmark  &  20.9   \\
\xmark& \cmark & \cmark & 22.7  \\ \hline
\cmark&\cmark&\cmark&\textbf{25.4}\\\hline
\end{tabular}}
\end{table}

\noindent \textbf{Analysis of the Local Relational Block:} We also dig deeper in to the Local Relational Block in each stage. As shown in Fig.~\ref{fig:stage}, there are two linear layers and one temporal convolutional layer in a Local Relational Block. In Table~\ref{table:local}, we further perform ablations of these components. 
First, we find that without the temporal convolutional layer, the detection performance drops. This observation shows the importance of mixing the transformer tokens with a temporal locality. 
Second, we study the importance of the transition layer (i.e., linear layer).  
When the feature size remains constant, having the transition layer can boost the performance by + 1.8\%, which shows the importance of such transition layers. 
Finally, we study how the expansion rate affects the network performance. While setting different feature expansion rates, we find that temporal convolution can better model the local temporal relations when the input feature is in a higher dimensional space.  
\begin{table}[t]\tabcolsep=12pt
\caption{\textbf{Ablation on the design of Local Relational Block:} Per-frame mAP on Charades using only RGB input. \xmark~ indicates we remove the linear or temporal convolutional layer. Feature expansion rate 1 indicates that the feature-size is not changed in the Local Relational Block.}
\label{table:local}
\centering
\scalebox{0.83}{
\begin{tabular}{cc|c}
\hline
Feature Expansion  & Temporal   & mAP \\
 Rate ($\theta$) &Convolution&(\%) \\ \hline
8     & \xmark     & 22.3    \\\hline
\xmark & \cmark & 22.4     \\
1     & \cmark & 24.2    \\\hline
4     & \cmark &  24.9   \\
8     & \cmark &   \textbf{25.4}  \\\hline
\end{tabular}}
\end{table}

\begin{table}[t!]\tabcolsep=2pt
\centering
\caption{\textbf{Comparison with the state-of-the-art methods} on three densely labelled datasets. 
Backbone indicates the visual encoder. Note that the evaluation for the methods is based on per-frame mAP (\%) using only RGB videos. }
\label{tab:SoTA}
\scalebox{0.82}{
\begin{tabular}{l|c|c|ccc}
\hline
            & Backbone & GFLOPs & Charades & MultiTHUMOS & TSU                  \\\hline
R-C3D~\cite{RC3d}       & C3D      &  -    & 12.7                 & -           & 8.7                  \\
Super-event~\cite{superevent} & I3D    &  0.8  & 18.6                 & 36.4        & 17.2                 \\
TGM~\cite{TGM1}         & I3D   &  1.2    & 20.6                 & 37.2        & 26.7                 \\
PDAN~\cite{Dai_2021_WACV}        & I3D  &   3.2    & 23.7                 & 40.2        & 32.7                 \\
Coarse-Fine~\cite{kahatapitiya2021coarse} & X3D &  -  & 25.1                 & -           & -                    \\
MLAD~\cite{MLAD}        & I3D & 44.8         & 18.4                 & 42.2        & -                    \\\hline\hline
\textbf{MS-TCT}      & I3D  &   6.6    & \textbf{25.4}        &  \textbf{43.1}  &\textbf{33.7}\\\hline
\end{tabular}}
\end{table}

\begin{table*}[h]
\centering
\caption{\textbf{Evaluation on the Charades dataset using the action-conditional metrics~\cite{MLAD}:} Similar to MLAD, both RGB and Optical flow are used for the evaluation. $P_{AC}$ - Action-Conditional Precision, $R_{AC}$ - Action-Conditional Recall, $F1_{AC}$ - Action-Conditional F1-Score, $mAP_{AC}$ - Action-Conditional Mean Average Precision. $\tau$ indicates the temporal window size.} 
\label{tab_cooccuring}
{
\scalebox{0.85}{
\begin{tabular}{l|cccc|cccc|cccc}
\hline
     & \multicolumn{4}{|c|}{$\tau$ = 0} & \multicolumn{4}{|c|}{$\tau$ = 20} & \multicolumn{4}{|c}{$\tau$ = 40} \\\hline
     & $P_{AC}$   & $R_{AC}$   & $F1_{AC}$   & $mAP_{AC}$  & $P_{AC}$   & $R_{AC}$   & $F1_{AC}$   & $mAP_{AC}$   & $P_{AC}$   & $R_{AC}$   & $F1_{AC}$   & $mAP_{AC}$   \\\hline
I3D  &14.3     &1.3     &2.1      &15.2      &12.7     &1.9     &2.9      &21.4       &14.9     &2.0     &3.1      &20.3       \\
CF   &10.3     &1.0     &1.6      &15.8      &9.0     &1.5     &2.2      &22.2       &10.7     &1.6     &2.4      &21.0       \\
MLAD~\cite{MLAD} &19.3     &7.2     &8.9      &28.9      &18.9     &8.9     &10.5      &35.7       &19.6     &9.0     &10.8      &34.8       \\\hline
\textbf{MS-TCT} &\textbf{26.3}     &\textbf{15.5}     &\textbf{19.5}     &\textbf{30.7}      &\textbf{27.6}     &\textbf{18.4}     &\textbf{22.1}      &\textbf{37.6}       &\textbf{27.9}     &\textbf{18.3}     &\textbf{22.1}      &\textbf{36.4}      \\\hline
\end{tabular}
}} 
\end{table*}

\subsection{Comparison to the State-of-the-Art }
In this section, we compare MS-TCT with the state-of-the-art action detection methods (see Table~\ref{tab:SoTA}). 
Proposal based methods, such as R-C3D~\cite{RC3d} fail in multi-label datasets due to the highly-overlapping action instances, which challenge the proposal and NMS-based methods. 
Superevent~\cite{superevent} superimposes a global representation to each local feature based on a series of learnable temporal filters. 
However, the distribution of actions varies from one video to the other. As super-event learns a fixed filter location for all the videos in the training distribution, this location is suitable to mainly actions
with high frequency. 
%
TGM~\cite{TGM1}
and PDAN~\cite{Dai_2021_WACV} are methods based on temporal convolution of video segments. 
Nevertheless, those methods only process videos locally at a single temporal scale. Thus, they are not effective in modeling long-term dependencies and high-level semantics. 
%
Coarse-Fine Network~\cite{kahatapitiya2021coarse} achieves 25.1\% on Charades. However, this method is built on top of the video encoder X3D~\cite{x3d}, which prevents the usage of higher number of input frames. Moreover, it relies on a large stride between the frames. Therefore, it fails to model fine-grained action relations, and can not process long videos in MultiTHUMOS and TSU.
MLAD~\cite{MLAD} jointly models action class relations for every time-step and temporal relations for every class. This design leads to a huge computational cost, while under-performing on datasets with a large number of action classes (e.g. Charades). 
Thanks to the combination of transformer and convolution in a multi-scale hierarchy, the proposed MS-TCT consistently outperforms previous state-of-the-art methods in all three challenging multi-label action detection datasets that we considered. 
We also compare the computational requirement (FLOPs) for the methods built on top of the same Visual Encoder (i.e., I3D features), taking as input the same batch of data. We observe that the FLOPs of MS-TCT is higher with a reasonable margin than pure convolutional methods (i.e., PDAN, TGM, super-event). 
However, compared to a transformer based action detection method MLAD, MS-TCT uses only $\frac{1}{7}$th of the FLOPs. 
%

We also evaluate our network with the action-conditional metrics introduced in~\cite{MLAD} on Charades dataset in Table~\ref{tab_cooccuring}. 
These metrics are used to measure a method’s ability to model both co-occurrence dependencies and temporal dependencies of action classes. 
Although our network is not specifically designed to model cross-class relations as in MLAD, it still achieves higher performance on all action-conditional metrics with a large margin, showing that MS-TCT effectively models action dependencies both within a time-step (i.e., co-occurring action, $\tau$ = 0) and throughout the temporal dimension ($\tau>$ 0).

\begin{figure}[t!]
\centering
\includegraphics[width=8cm]{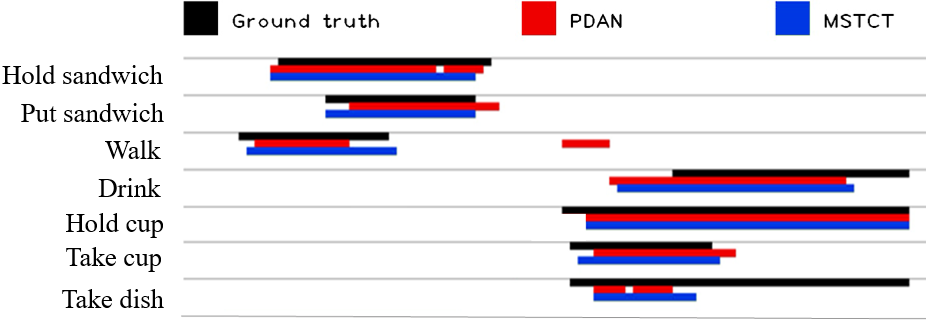}\vspace{-0.1in}
\caption{\textbf{Visualization of the detection results} on an example video  along time axis. In this figure, we visualize the ground truth and the detection of PDAN and MS-TCT. 
}
\label{fig:detection_visual}
\end{figure}

Finally, we present a qualitative evaluation for PDAN and MS-TCT on the Charades dataset in Fig.~\ref{fig:detection_visual}. As the prediction of the Coarse-Fine Network is similar to the X3D network which is limited to dozens of frames, thus we can not compare with the Coarse-Fine network on the whole video. Here, we observe that MS-TCT can predict action instances more precisely compared to PDAN. This comparison reflects the effectiveness of the transformer architecture and multi-scale temporal modeling.

\begin{table}[htp]
\begin{floatrow}
\floatsetup{floatrowsep=qquad,captionskip=1pt} \tabcolsep=9pt
\begin{floatrow}
\hspace{-0.7cm}
\scalebox{1}{
\ttabbox{\caption{\small \textbf{Study on stage type} showing the effect of having both convolutions and self-attention. 
}}{
\centering
\scalebox{0.9}{
\label{tab:type}
\begin{tabular}{l|c}
\hline
Stage-Type               & mAP   \\\hline
Pure Transformer         & 22.3  \\
Pure Convolution         & 21.4  \\
ConvTransformer &\textbf{25.4}   \\\hline
\end{tabular}}}}
\end{floatrow}

\hspace{-0.6cm}
\renewcommand\arraystretch{1.0} 
\scalebox{1}{
\floatsetup{floatrowsep=qquad,captionskip=1pt} \tabcolsep=7pt
\begin{floatrow}
\ttabbox{\caption{\small \textbf{Study on $\mathbf{\sigma}$} showing the effect of scale of Gaussians in heat-maps.  
}}{
\centering
\scalebox{0.9}{
\label{Tab:sigma}
\begin{tabular}{l|c}
\hline
Variance: $\sigma$ & $\;$mAP$\;$   \\\hline
1/8 duration &  24.6     \\
1/4 duration &  24.8     \\
1/2 duration &  \textbf{25.4}     \\\hline
\end{tabular}}}
\end{floatrow}}
\end{floatrow}
\end{table}

\subsection{Discussion and Analysis}

\noindent \textbf{Transformer, Convolution or ConvTransformer?}
To confirm the effectiveness of our ConvTransformer, we compare with a pure transformer network and a pure convolution network. Each network has the same number of stages as MS-TCT with similar settings (e.g. blocks, feature dimension).
%
%
In pure transformer, a pooling layer and a linear layer constitute the temporal merging block, followed by $B$ transformer blocks in each stage. A transformer block is composed of a multi-head attention layer, norm-add operations and a feed-forward layer. A learned positional embedding is added to the input tokens to encode the positional information. This pure transformer architecture achieves 22.3\% on Charades.
In pure convolution-based model, we retain the same temporal merging block as in MS-TCT, followed by a stack of $B$ temporal convolution blocks. Each block consists of a temporal convolution layer with a kernel-size of $k$, a linear layer, a non-linear activation and a residual link. This pure temporal convolution architecture achieves 21.4\% on Charades.  
%
In contrast, the proposed ConvTransformer outperforms both the pure transformer and the pure convolutional network by a large margin (+ 3.1\%, and + 4.0\% on Charades, respectively. See Table~\ref{tab:type}).
It shows that ConvTransformer can better model the temporal relations of complex actions. 

\begin{figure}[t!]
\centering
\includegraphics[width=6.7cm]{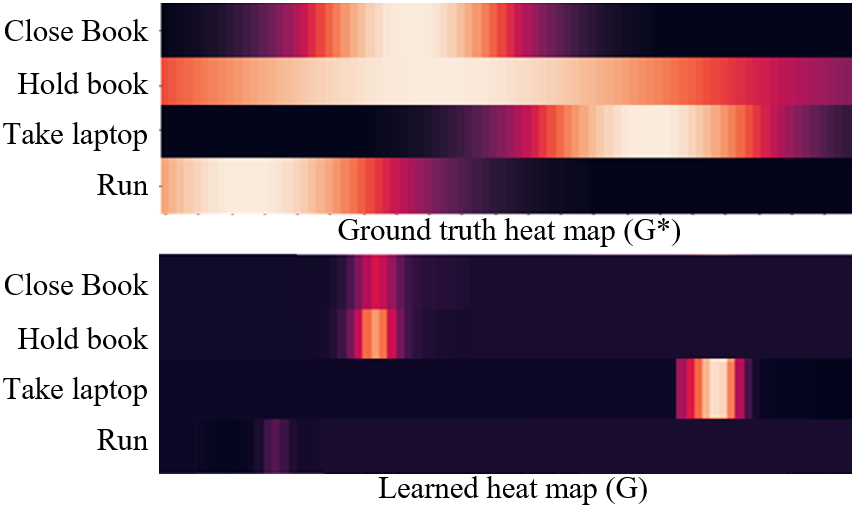} 
\caption{\textbf{Heat-map visualization along time axis:} On the top, we show the ground truth heat-map ($G^*$) of the example video. On the bottom is the corresponding learned heat-map ($G$) of MS-TCT. As the heat-map is generated by a Gaussian function, the lighter region indicates closer to the center of the instance. 
}
\label{fig:heatmap}
\end{figure}

\noindent\textbf{Heat-map Analysis:} We visualize the ground truth heat-map ($G^*$) and the corresponding predicted heat-map ($G$) in Fig.~\ref{fig:heatmap}. We observe that with the heat-map branch, MS-TCT predicts the center location of the action instances, showing that MS-TCT embeds the center-relative information in to the tokens. However, as we optimize with the focal loss to highlight the center, the boundaries of the action instance in this heat-map are less visible. 
%
%
We then study the impact of $\sigma$ on performance. As shown in Table~\ref{Tab:sigma}, we set $\sigma$ to be either $\frac{1}{8}$, $\frac{1}{4}$ or $\frac{1}{2}$ of the instance duration while generating the ground-truth heat-map $G^*$. MS-TCT improves by + 0.5\%, + 0.7\%, + 1.3\% respectively \wrt the MS-TCT without the heat-map branch, when $G^*$ set to different $\sigma$. This result reflects that a larger $\sigma$ can better provide the center-relative position. 
We investigate further by adding a heat-map branch to another action detection model: PDAN~\cite{Dai_2021_WACV}. 
Although the heat-map branch also improves PDAN (+ 0.4 \%), the relative improvement is lower compared to MS-TCT (+ 1.3 \%).
Our method features a multi-stage hierarchy along with a TS Mixer. As the heat-map branch takes input from all the stages, the center-relative position is embedded even in an early stage. Such tokens with the relative position information, when fed through the following stages, benefits the multi-head attention to better model temporal relations among the tokens. This design makes MS-TCT to better leverage the heat-map branch compared to PDAN. 
%

\noindent\textbf{Temporal Positional Embedding:}
We further study whether the Temporal Encoder of MS-TCT benefits from positional embedding. 
We find that the performance drops by 0.2\% on Charades when a learnable positional embedding~\cite{dosovitskiy2020image} is added to the input tokens before processing them with the Temporal Encoder. This shows that the current design can implicitly provide a temporal positioning for the tokens. Adding further positional information to the tokens makes it redundant, leading to lower detection performance. 

\vspace{0.1in}
\section{Conclusion}
\label{sec:conclusion}
In this work, we proposed a novel ConvTransformer network: MS-TCT for action detection. It benefits from both convolutions and self-attention to model local and global temporal relations respectively, at multiple temporal scales. Also, we introduced a new branch to learn class-wise relative positions of the action instance center. MS-TCT is evaluated on three challenging densely-labelled action detection benchmarks, on which it achieves new state-of-the-art results.

\vspace{0.1in}
\noindent\textbf{Acknowledgements:} 
This work was supported by the French government, through the 3IA Côte d’Azur Investments in the Future project managed by the National Research Agency with the reference number ANR-19-P3IA-0002. This work was also supported in part by the National Science Foundation (IIS-2104404 and CNS-2104416). The authors are grateful to the OPAL infrastructure from Université Côte d'Azur for providing resources and support.


{\small
\bibliographystyle{ieee_fullname}
\bibliography{egbib}
}
\newpage 
~
\newpage 
\noindent\textbf{\LARGE{Appendix}} 
\vspace{0.1in}

\noindent In the following sections, we provide further experimental results on MS-TCT along four aspects: (1) temporal action relation, (2) blurred videos, (3) heat-map branch (4) hyper-parameters. In addition, we provide the limitation of our method and more details on the Temporal Encoder architecture.
%

\vspace{0.1in}
\noindent \textbf{\large{A1. Analysis of Temporal Action Relations}}
\vspace{0.1in}

Firstly, we analyse how different types of layers in the stages affect the range of temporal relations. 
We utilize the action-conditional metrics~\cite{MLAD} for this analysis as it provides the dependencies between the video tokens at different temporal ranges. 
Similar to Section~4.3 in the main paper, we construct three types of stage based on the temporal encoder: Pure Convolution, Pure Transformer and ConvTransformer (i.e., MS-TCT). 
As shown in table~\ref{tab:dependencies}, we find that Pure Convolution is better than Pure Transformer for local temporal dependencies ($\tau = 5$), but for the long-term dependencies ($\tau =100$), the Transformer based model achieves better performance. 
The ConvTransformer benefits from both layers thus achieving better performance on both short and long term dependencies.

\begin{table}[h]
\caption{Studies on temporal dependencies. Evaluated with action conditional mAP~\cite{MLAD} on Charades using RGB.}
\label{tab:dependencies}
\begin{tabular}{l|cc}
\hline
Stage-Type       & $\tau =5$  & $\tau =100$  \\\hline
Pure Convolution & 26.4 & 28.7 \\
Pure Transformer  & 24.6 & 30.8 \\
ConvTransformer  & \textbf{28.9} & \textbf{33.1}    \\ \hline
\end{tabular}
\end{table}

\vspace{0.1in}
\noindent \textbf{\large{A2. Performance on Blurred Videos}} 
\vspace{0.1in}

For the protection of personal privacy, the face of all the subjects is blurred on TSU~\cite{dai2020toyota}. The proposed MS-TCT outperforms the state-of-the-art methods on this dataset, showing that MS-TCT is not relying on the information of person id to conduct the action detection. 

\vspace{0.1in}
\noindent \textbf{\large{A3. More studies on Heat-map Branch}}
\vspace{0.1in}

We first study how the location of the heat-map branch affects the detection performance. 
Precisely, instead of feeding the output features from all stages (i.e., MS-TCT), here, we provide only the stage 1 or stage 4 features to the heat-map branch. 
%
In table~\ref{tab:location}, we find that having the heat-map branch either in the early stage (i.e., stage $=1$) or at late stage (i.e., stage $=4$) can boost the performance of MS-TCT compared to MS-TCT without the heat-map branch.
The model achieves the overall best performance, while exploiting features from all the stages (i.e., $F_v$) to the heat-map branch. 

\begin{table}[h]
\caption{Location of heat-map branch on Charades dataset using only RGB. Stage indicates the features from which stage is fed to the heat-map branch. \xmark indicates not having the heat-map branch and $All$ indicates that we fed features from all the stages to the heat-map branch (i.e., similar to MS-TCT).  }
\label{tab:location}
\begin{tabular}{c|c}
\hline
Stage & mAP (\%) \\\hline
\xmark & 24.1 \\
1     & 24.9 \\
4     & 24.7 \\
All & \textbf{25.4} \\\hline
\end{tabular}
\end{table}

\begin{figure}[h!]
\centering
\includegraphics[width=7.5cm]{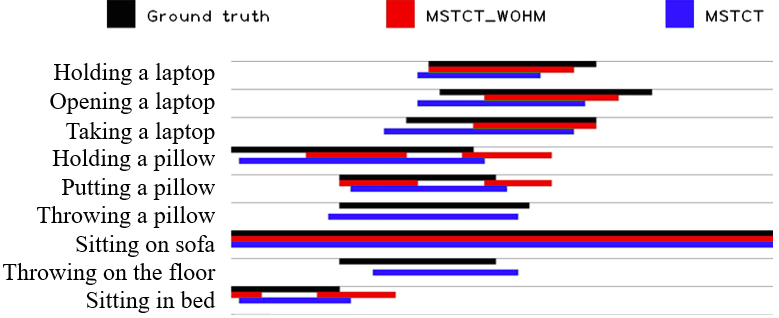}
\caption{Qualitative study for the heat-map branch. Blue: the proposed method MS-TCT. Red: MS-TCT without the heat-map branch. 
}
\label{fig:heatmap_qualitative}
\end{figure}

We then perform a qualitative analysis of the action detection performance for MS-TCT, with or without the heat-map branch. As shown in figure~\ref{fig:heatmap_qualitative}, we find that while having the heat-map branch (i.e., MS-TCT), the prediction is more continuous (e.g., \textit{sitting in bed}, \textit{putting a pillow}). This reflects that with the heat-map branch, the tokens in MS-TCT are embedded with the instance-center relative position. Therefore, the tokens in the instance, especially the ones close to the center region, are well detected. 

\vspace{0.1in}
\noindent \textbf{\large{A4. More Studies on Hyper-parameters}}
\vspace{0.1in}

In this section, we further study the hyper-parameters of MS-TCT model on the Charades dataset. 

\noindent\textbf{Study on the number of heads $H$.}
Multi-head attention layer divides the channels into several groups. Each group of features is sent to an attention head to model the global temporal relation. 
While changing the number of heads, we find that the FLOPs number remains the same, thanks to the group operations. With more heads, more complex relationships can be modelled. However, increasing the heads reduces the number of channels processed by each head. 
As a balance between the number of channels for each head and the number of relations to model, we set the number of heads $H$ to 8. 

\begin{table}[h]
\caption{Study on number of heads in Global Relational Blocks on Charades dataset using only RGB. }
\label{tab:head}
\begin{tabular}{cc|c}
\hline
\# Heads $H$ & GFLOPs & mAP (\%) \\\hline
1     &6.6 & 24.6 \\
4     &6.6 & 25.1 \\
8     &6.6 & \textbf{25.4} \\
16    &6.6 & 25.3 \\\hline
\end{tabular}
\end{table}

\noindent\textbf{Study on the number of Blocks $B$.}
We then study how the number of Global-Local Relational Blocks $B$ affects the network performance. From table~\ref{tab:block}, we find that the FLOPs number is increasing with the number of blocks. The network can achieve better performance when more Global and Local layers are involved for the temporal modeling.
As a balance between FLOPs and the number of blocks, we utilize a three-block architecture. 

\begin{table}[t]
\caption{Study on number of Global-Local Relational Blocks for each stage on Charades dataset using only RGB.}
\label{tab:block}
\begin{tabular}{cc|c}
\hline
\# Block $B$ & GFLOPs & mAP (\%)      \\\hline
1       &3.4 & 24.3    \\
2       &5.0 & 24.7   \\
3       &6.6 & 25.4      \\
4       &8.2 & 25.5      \\\hline
\end{tabular}
\end{table}

\noindent\textbf{Study on the kernel size $K$ for Temporal Convolution.}
After that, we study how the kernel size of the temporal convolution in the Local Relational Block affects the action detection performance. 
In table~\ref{tab:kernel}, we find that removing the temporal convolution layer in the Local Relational Block causes a significant drop in the performance. 
While having the convolution layer, there is not a large difference between the model with different kernel sizes. As a larger kernel size results in more weight parameters, in this work we choose the kernel size as $3$. 

\noindent\textbf{Number of Tokens $T$.}
We randomly select consecutive $T$ tokens for each video in the training phase and utilize the sliding window at inference. 
Here, we have studied how the number of tokens $T$ affects the action detection performance. 
When $T$ is set to 128, 256 and 512 tokens, MS-TCT achieves 25.0\%, 25.4\% and 25.5\% on Charades. There is no significant difference in the action detection performance while changing the number of input tokens. However, increasing the number of tokens $T$ in MS-TCT increases the FLOPs. For the trade-off between the computation cost and performance precision, we set $T$ to 256 tokens, which corresponds to 2048 frames (about~86 sec.) of video.

\begin{table}[t]
\centering
\caption{Study on the kernel size $K$ for the temporal convolutional layer in Local Relational Block on Charades dataset using only RGB. \xmark indicates removing the temporal convolution layer in the Local Relational Block. }
\label{tab:kernel}
\begin{tabular}{c|c}
\hline
\# Kernel Size $K$ & mAP (\%)    \\\hline
\xmark & 22.3 \\
3       & 25.4    \\
5       & 25.1   \\
7       & 25.4  \\\hline
\end{tabular}
\end{table}

\noindent\textbf{Study on the kernel size $N$ for Temporal Encoder.}
Finally, we analyse the hyper-parameter $N$ in table~\ref{tab:numofstage}.
Number of Stages (N) determines the level of semantic information in our representations. Our experiment show that a 4-stage structure strikes a balance between the performance and model size. 

\begin{table}[t!]
\centering
\caption{Study on the number of stage $N$ for the temporal encoder. }
\label{tab:numofstage}
\begin{tabular}{c|ccccc}
\hline
$N$ (\#Stage) & 1& 2 & 3 & 4 & 5  \\\hline
Charades mAP & 20.4 & 22.9  & 24.6  & 25.4  & 25.6   \\\hline
\end{tabular}
\end{table}

\vspace{0.1in}
\noindent \textbf{\large{A5. Method Limitation}}
\vspace{0.1in}

Although MS-TCT has outperformed state-of-the-art methods on three challenging datasets, the performance is still relatively low (e.g., less than 30\% on Charades). One of the reasons is that the Visual Encoder and the Temporal Encoder in MS-TCT are not optimized jointly in our network, due to hardware limitation. Our future work will focus on modeling the temporal and spatial relations end-to-end for long untrimmed videos. 

\vspace{0.1in}
\noindent \textbf{\large{A6. Which actions benefit the most?}}
\vspace{0.1in}

\begin{figure}[h!]
\centering
\includegraphics[width=7.5cm]{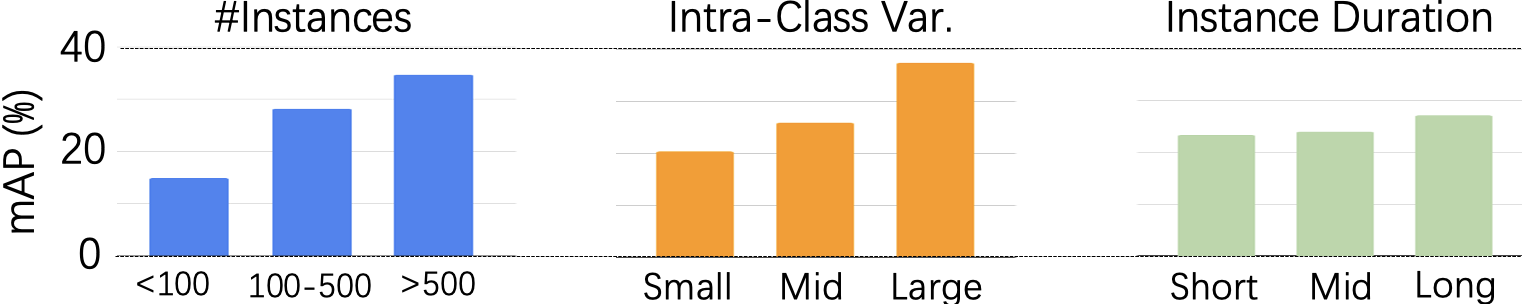}
\caption{The sensitivity of MS-TCT's mAP to three action characteristics. 
}
\label{fig:comparaison_class}
\end{figure}

In order to quantify the action types which are the most benefited from MS-TCT, we present performance w.r.t. 3 action-class characteristics (Fig.~\ref{fig:comparaison_class}): \#~instances, intra-class variance of duration and normalized instance duration~\cite{alwassel2018diagnosing}. Note that, we normalize the length of the instance by the duration of the video to have the normalized instance duration. 
Firstly, we find that as MS-TCT does not have a specific design for imbalanced data, this model is troubled in few-sampled action classes. 
Secondly, we notice that MS-TCT can perform better in the action class with high intra-class variance of duration. 
Finally, by the analysis of action classes with different instance duration, we find that MS-TCT can consistently detect both long and short instances. 

\noindent \textbf{\large{A7. Temporal Encoder Architecture}}
\label{sec:srchitecture}
\vspace{0.1in}

To better understand the computation flow, table~\ref{tab:architecture} shows the detailed architecture along with the input and output feature size of our Temporal Encoder Module. 
We have also allocated different hyper-parameters as $H$, $B$ for different stages. However, we do not observe further improvements. 

\begin{table*}[h!] 
\caption{Temporal Encoder architecture. The input and out feature size is following $T \times D$ format, where number of tokens ($T$) on the left and feature dimension $D$ on the right. Linear layer is the kernel size 1 convolution. For the hyper-parameters: H: heads, K: kernel size, S: stride, P: zero-padding rate. Note that, for brevity, number of blocks ($B$) is not reflected in this table. Each Stage contains 3 Global-Local Relational Blocks, i.e., the set of a Global Relational Block and a Local Relational Block repeated 3 times for each stage. }
\label{tab:architecture}
\begin{tabular}{c|l|l|c|c|c}
\hline
Stage                                         & Components                      & Learnable layers     & Hyper-parameters             & Input size         & Output size        \\ \hline
\multicolumn{1}{c|}{\multirow{5}{*}{Stage 1}} & Temporal Merge                      & Temporal Convolution           & K: 3, S: 1, P: 1 & 256$\times$1024 & 256$\times$256  \\ \cline{2-6} 
\multicolumn{1}{c|}{}                         & Global Relation                 & Multi-head Self-Attention & H: 8             & 256$\times$256  & 256$\times$256  \\ \cline{2-6} 
\multicolumn{1}{c|}{}                         & \multirow{3}{*}{Local Relation} & Linear               & K: 1, S: 1, P: 0 & 256$\times$256  & 256$\times$2048 \\ 
\multicolumn{1}{c|}{}                         &                                 & Temporal Convolution           & K: 3, S: 1, P: 1 & 256$\times$2048 & 256$\times$2048 \\ 
\multicolumn{1}{c|}{}                         &                                 & Linear               & K: 1, S: 1, P: 0 & 256$\times$2048 & 256$\times$256  \\ \hline
\multirow{5}{*}{Stage 2}                      & Temporal Merge                      & Temporal Convolution           & K: 3, S: 2, P: 1 & 256$\times$256  & 128$\times$384  \\ \cline{2-6} 
                                              & Global Relation                 & Multi-head Self-Attention & H: 8             & 128$\times$384  & 128$\times$384  \\ \cline{2-6} 
                                              & \multirow{3}{*}{Local Relation} & Linear               & K: 1, S: 1, P: 0 & 128$\times$384  & 128$\times$3072 \\ 
                                              &                                 & Temporal Convolution           & K: 3, S: 1, P: 1 & 128$\times$3072 & 128$\times$3072 \\ 
                                              &                                 & Linear               & K: 1, S: 1, P: 0 & 128$\times$3072 & 128$\times$384  \\ \hline
\multirow{5}{*}{Stage 3}                      & Temporal Merge                      & Temporal Convolution           & K: 3, S: 2, P: 1 & 128$\times$384  & 64$\times$576   \\ \cline{2-6} 
                                              & Global Relation                 & Multi-head Self-Attention & H: 8             & 64$\times$576   & 64$\times$576   \\ \cline{2-6} 
                                              & \multirow{3}{*}{Local Relation} & Linear               & K: 1, S: 1, P: 0 & 64$\times$576   & 64$\times$4608  \\ 
                                              &                                 & Temporal Convolution           & K: 3, S: 1, P: 1 & 64$\times$4608  & 64$\times$4608  \\ 
                                              &                                 & Linear               & K: 1, S: 1, P: 0 & 64$\times$4608  & 64$\times$576   \\ \hline
\multirow{5}{*}{Stage 4}                      & Temporal Merge                      & Temporal Convolution           & K: 3, S: 2, P: 1 & 64$\times$576   & 32$\times$864   \\ \cline{2-6} 
                                              & Global Relation                 & Multi-head Self-Attention & H: 8             & 32$\times$864   & 32$\times$864   \\ \cline{2-6} 
                                              & \multirow{3}{*}{Local Relation} & Linear               & K: 1, S: 1, P: 0 & 32$\times$864   & 32$\times$6912  \\
                                              &                                 & Temporal Convolution           & K: 3, S: 1, P: 1 & 32$\times$6912  & 32$\times$6912  \\
                                              &                                 & Linear               & K: 1, S: 1, P: 0 & 32$\times$6912  & 32$\times$864  \\ \hline
\end{tabular}
\end{table*}

\end{document}